\documentclass{article}

\usepackage[main, preprint]{neurips_2025}

\usepackage[utf8]{inputenc} 
\usepackage[T1]{fontenc}    
\usepackage{microtype}      

\usepackage{hyperref}       
\usepackage{url}            

\usepackage{amsfonts}       
\usepackage{amsmath}
\usepackage{amssymb}        
\usepackage{nicefrac}       

\usepackage{xcolor}         
\usepackage{graphicx}
\usepackage{subcaption}
\usepackage{float}

\usepackage{booktabs}       

\title{KAN-Dreamer: Benchmarking Kolmogorov-Arnold Networks as Function Approximators in World Models}

\author{%
  Chenwei Shi \\
  Shanghai Research Institute for Intelligent Autonomous Systems \\
  Tongji University \\
  Shanghai 200092, China \\
  \texttt{2511973@tongji.edu.cn} \\
  \And
  Xueyu Luan \\
  School of Electronic and Information Engineering \\
  Tongji University \\
  Shanghai 200092, China \\
  \texttt{2531946@tongji.edu.cn} \\
}

\begin{document}

\maketitle

\begin{abstract}
  DreamerV3 stands as a state-of-the-art \textbf{online} model-based reinforcement learning (MBRL) algorithm, achieving remarkable sample efficiency across diverse control tasks by learning behaviors within a latent world model. Concurrently, Kolmogorov-Arnold Networks (KANs) have emerged as a promising alternative to traditional Multi-Layer Perceptrons (MLPs). Rooted in the Kolmogorov-Arnold representation theorem, KANs demonstrate superior parameter efficiency and interpretability compared to fixed activation architectures. To mitigate the computational overhead of original KANs, variants such as FastKAN leverage Radial Basis Functions (RBFs) to accelerate inference while maintaining approximation capabilities. In this work, we investigate the feasibility and potential advantages of integrating KAN architectures into the DreamerV3 framework. We introduce \textbf{KAN-Dreamer}, a modified architecture where specific MLP and convolutional components of the original DreamerV3 are replaced by KAN and FastKAN layers. \textbf{To ensure computational efficiency and compatibility within the JAX-based World Model, we implement a tailored, fully vectorized version of these networks with simplified grid management.} \textbf{Specifically, we structure our investigation into three functional subsystems: Visual Perception (Encoder/Decoder), Latent Prediction (Reward/Continue), and Behavior Learning (Actor/Critic).} We conduct empirical evaluations on the \texttt{walker\_walk} task from the DeepMind Control Suite (DMC). Our study analyzes the impact of these architectural changes on training dynamics, specifically focusing on sample efficiency, wall-clock training time, and asymptotic performance. \textbf{Experimental results demonstrate that utilizing our adapted FastKAN as a drop-in replacement for the Reward and Continue predictors yields performance on par with the original MLP-based architecture, maintaining parity in both sample efficiency and training speed.} This report serves as a preliminary study for future developments in KAN-based world models.
\end{abstract}

\section{Introduction}
\label{Introduction}

World models have emerged as a pivotal paradigm in the pursuit of Artificial General Intelligence (AGI) and Embodied AI, aiming to learn implicit representations of the external world and predict future dynamics. This concept has found profound applications across various domains, ranging from video generation (e.g., Sora \citep{openai2024sora}, Veo \citep{deepmind2024veo}) and world simulators (e.g., Genie series \citep{bruce2024genie}) to autonomous driving systems (e.g., FSD V12 \citep{elluswamy2023tesla}, GAIA-1 \citep{hu2023gaia}, UniAD \citep{hu2023uniad}). Particularly in the realm of Embodied AI and robotics, world models offer a critical solution to the challenge of sample efficiency by serving as the core of Model-Based Reinforcement Learning (MBRL). The Dreamer algorithm series \citep{hafner2019dream, hafner2020mastering, hafner2025mastering, hafner2025training}, along with its application to physical robots \citep{wu2023daydreamer}, stands as a cornerstone of modern MBRL. By leveraging real interaction data to train a world model and subsequently performing "imagination-based" training within latent trajectories, Dreamer achieves remarkable sample efficiency compared to Model-Free approaches, especially in scenarios where real-world interactions are costly or scarce. While DreamerV3 \citep{hafner2025mastering} has demonstrated state-of-the-art performance across diverse control tasks, and the recent DreamerV4 \citep{hafner2025training} has introduced Diffusion Transformers to address long-horizon prediction errors, the exploration of efficient architectural backbones within the Dreamer framework remains an open avenue of research.

Currently, DreamerV3 relies on a hybrid architecture, utilizing Multi-Layer Perceptrons (MLPs) for latent dynamics, latent prediction and behavior learning components (e.g., Reward Predictor, Dynamics Predictor, Actor, and Critic), while employing Convolutional Neural Networks (CNNs) for processing visual observations. Recently, Kolmogorov-Arnold Networks (KANs) \citep{liu2024kan} have entered the spotlight as a promising alternative to these traditional neural building blocks. Grounded in the Kolmogorov-Arnold representation theorem, KANs place learnable activation functions on the edges of the network rather than the nodes. This structural innovation offers theoretical advantages, including superior function approximation capabilities, higher parameter efficiency, and enhanced interpretability. Furthermore, KANs have rapidly emerged as a powerful paradigm for `AI for Science,' bridging the gap between connectionist learning and scientific discovery \citep{liu2024kan2}. Although original KAN implementations suffer from slower inference speeds, recent variants such as FastKAN \citep{li2024kolmogorov} have mitigated these computational bottlenecks by utilizing techniques like Radial Basis Functions (RBFs), making them viable for large-scale integration.

In this technical report, we present an exploratory study integrating KANs and their efficient variant, FastKAN, into the DreamerV3 algorithm. We attempt to replace the conventional MLP and CNN structures within DreamerV3 to assess the feasibility and potential benefits of KAN-based world models. Our contributions are as follows:

\begin{itemize}
    \item We implement KAN and FastKAN within the DreamerV3 framework, providing the first empirical evidence of their viability as replacements for standard MLPs in complex latent dynamics models.
    \item We critically analyze the trade-offs between the superior fitting capabilities of KANs and their inference latency, investigating their impact on the generation speed of imagination trajectories and the overall sample efficiency of the agent.
\end{itemize}

The rest of this paper is structured as follows: Section \ref{Related Work} and Section \ref{Preliminaries} review related work and necessary preliminaries, respectively. Section \ref{Method} details the architecture of KAN-Dreamer and our implementation strategy. Section \ref{Experiments} presents our experimental results on standard benchmarks, followed by a discussion in Section \ref{Discussion}. Finally, Section \ref{Conclusion} concludes the work.

\section{Related Work}
\label{Related Work}

\subsection{World Models and the Dreamer Family}
World models aim to learn compact latent representations that capture the underlying structure and dynamics of an environment, enabling agents to reason, plan, and learn policies through imagination rather than direct interaction. The seminal work World Models \citep{ha2018world} first demonstrated that a generative latent dynamics model can support training controllers inside a learned “dream,” establishing the feasibility of imagination-based reinforcement learning (RL). Later, PlaNet \citep{hafner2019learning} improved multi-step latent consistency by introducing the Recurrent State-Space Model (RSSM), which combines deterministic and stochastic latent components to enhance long-horizon predictive accuracy. This modeling strategy became foundational for the Dreamer family.

DreamerV1 (Dream to Control) \citep{hafner2019dream} showed that agents can learn policies and value functions entirely from imagined trajectories generated by a latent world model. Building on this, DreamerV2 (Mastering Atari with Discrete World Models) \citep{hafner2020mastering} introduced improved training objectives, discrete latent representations, and more stable optimization, achieving strong performance across a wide range of visual-control tasks. Culminating this line of research, DreamerV3 (Mastering diverse control tasks through world models) \citep{hafner2025mastering} further established world models as a general-purpose and highly stable \textbf{online} RL paradigm by introducing normalization and balanced training mechanisms that enable robust performance across over 150 heterogeneous tasks. Notably, the most recent iteration, DreamerV4 (Training Agents Inside of Scalable World Models) \citep{hafner2025training}, pivoted this paradigm towards the \textbf{offline setting}. It replaces the Recurrent State-Space Model with an efficient transformer-based architecture, enabling high-fidelity video prediction and solving complex long-horizon tasks purely from offline data without any environment interaction.

Despite these algorithmic advancements, the underlying function approximators within the established RSSM-based Dreamer family (V1-V3) have remained predominantly based on Multi-Layer Perceptrons (MLPs) and Convolutional Neural Networks (CNNs). This long-standing reliance on traditional architectures motivates our investigation. As our work focuses on the \textbf{online} learning context, we use DreamerV3 as the foundational framework to explore whether emerging neural paradigms, specifically Kolmogorov-Arnold Networks, can serve as \textbf{competitive alternatives}.

\subsection{Kolmogorov-Arnold Networks and Efficient Variants}
The Kolmogorov–Arnold representation theorem establishes that any multivariate continuous function can be represented as a finite superposition of univariate continuous functions and addition operations \citep{kolmogorov1957representations, schmidt2021kolmogorov}. Motivated by this theorem, Kolmogorov–Arnold Networks (KANs) were proposed as a new neural architecture: unlike conventional multilayer perceptrons (MLPs) whose “weights” are scalar linear coefficients, KAN replaces each weight with a learnable univariate function, typically parameterized by splines \citep{liu2024kan}.

In the original KAN formulation, authors demonstrated that such networks — even with \textbf{significantly fewer neurons (units) and parameters} — can achieve comparable or better performance than larger MLPs on small-scale AI-for-science tasks, including function fitting and solving partial differential equations (PDEs). Moreover, KANs manifest “faster neural scaling laws” than MLPs in those settings. The use of edge-wise learnable functions also provides improved interpretability compared to black-box MLPs: one can inspect and visualize the learned univariate functions per edge. 

Nevertheless, the original KAN design faces practical challenges. Evaluating spline-based univariate functions on every connection incurs nontrivial computational and memory cost, leading to slower training and inference compared to MLPs. To address these drawbacks, \textbf{EfficientKAN} \citep{blealtan2024efficientkan} was proposed to accelerate computation by reformulating B-spline evaluations into matrix operations, significantly reducing memory overhead. Building on this momentum, \textbf{FastKAN} \citep{li2024kolmogorov} further simplified the architecture by replacing the complex B-spline basis with Gaussian Radial Basis Functions (RBFs). This modification not only retains the approximation capabilities but also offers a remarkably simple and implementation-friendly structure. The ease of vectorization and reproduction makes FastKAN particularly suitable for integration into complex systems like world models, motivating its adoption in our work.

Beyond the efficiency-focused optimizations of FastKAN, other research directions explore structural diversity, often by creating hybrid architectures that integrate MLP-style functions (e.g., KKAN \citep{toscano2025kkans}) or use alternative bases like sinusoids \citep{gleyzer2025sinusoidal}. While promising, these variants often introduce greater architectural complexity and more intricate implementation details. Consequently, to maintain a strictly controlled experimental setting focused on the core efficacy of KANs without confounding factors from hybrid designs, we prioritize the structurally simpler FastKAN for our world model integration.

However, despite these architectural optimizations, the capability of KANs—and \textbf{particularly their efficient variants like FastKAN}—to handle \textbf{complex control tasks} remains unproven. While KANs have begun to show promise in domains beyond toy regression, such as genomics and bioinformatics \citep{10.1093/bib/bbaf129}, their application as function approximators within the challenging context of \textbf{World Models} has not been comprehensively evaluated. To the best of our knowledge, no existing work has benchmarked KANs for the core components of a modern world model, such as for \textbf{high-dimensional visual reconstruction}, \textbf{latent-space reward prediction}, or \textbf{imagination-based policy learning}. This gap motivates our present work: integrating efficient KAN-based architectures into a world-model framework (in the spirit of Dreamer) and systematically evaluating their effectiveness.

\subsection{KANs in Reinforcement Learning}

Kolmogorov–Arnold Networks (KANs) have recently been explored as an alternative to MLPs for function approximation in reinforcement learning. Existing studies focus almost exclusively on model-free RL, inserting KANs into standard baselines such as PPO, SAC, A2C, and DQN/Rainbow to replace policy or value networks. These works report that KANs can match or surpass MLP agents with fewer parameters \citep{guo2025kan}, improve optimization smoothness \citep{vaca2024kolmogorov}, and offer interpretable univariate edge functions \citep{singh2025interpretable}.

However, all prior applications remain limited to direct policy/value learning, and no existing work studies KANs within Model-Based RL, specifically for the distinct challenges of \textbf{generative reconstruction} or \textbf{latent state interpretation}. This gap is significant because world models such as PlaNet and Dreamer rely heavily on the fidelity of these dense prediction heads to provide accurate learning signals for the agent \citep{hafner2019learning, hafner2025mastering, ha2018world}. Whether KANs' adaptive univariate functions can yield more parameter-efficient latent predictors or better visual decoders compared to standard architectures remains unexplored. Our work addresses this gap by providing the first systematic benchmark of KANs inside a world-model architecture, integrating them into the DreamerV3 pipeline to evaluate their suitability for visual representation learning, latent prediction, and imagination-based policy optimization.

\subsection{KANs in Generative Modeling and Representation Learning}

Kolmogorov–Arnold Networks (KANs) have recently attracted attention beyond regression, extending into generative modeling and representation learning. The architecture’s learnable univariate functions offer a potential advantage for representing complex conditional maps (e.g., latent $\rightarrow$ observation) with fewer parameters than traditional MLPs \citep{liu2024kan}. 

In the realm of generative modeling, KANs have been integrated into mainstream frameworks such as Diffusion Models and GANs. Recent works have replaced MLP or bottleneck layers in U-Nets with KAN blocks, showing competitive fidelity on specific tasks like medical image synthesis \citep{su2025kan}, low-light enhancement \citep{yeh2025diffusion}, and 3D scientific structure reconstruction \citep{wang2025ka}. These studies establish that KANs possess the capacity to model complex data distributions, a prerequisite for any world model.

For representation learning, the most relevant development to our work is the \textbf{Kolmogorov-Arnold Auto-Encoder (KAE)} \citep{yu2024kae}. Yu et al. demonstrated that KANs can effectively replace standard convolutional or MLP layers in autoencoders to reconstruct images. However, this evaluation was largely confined to simple, low-dimensional datasets such as MNIST. It remains an open question whether KAN-based encoders and decoders can scale to the high-dimensional, visually complex, and partially observable environments typical of modern Reinforcement Learning benchmarks (e.g., DeepMind Control Suite).

Furthermore, independent evaluations have highlighted potential scalability issues. Studies in computational mechanics and scientific computing report that while parameter-efficient, KANs can suffer from training instability and slower wall-clock convergence compared to highly optimized baselines \citep{wang2025kolmogorov,zhou2025askan}. It remains unverified whether these optimization difficulties persist or worsen when applied to the \textbf{high-dimensional, non-stationary visual data} inherent in World Models.

This context sets the stage for our investigation. DreamerV3 relies heavily on a robust \textbf{autoencoder structure (Visual Encoder and Decoder)} to extract compact representations from pixels. While KAE provides a proof-of-concept for simple static images, our work bridges the gap by benchmarking KAN architectures (specifically efficient variants) as \textbf{visual representation learners} within the rigorous and computationally demanding framework of World Models.

\section{Preliminaries}
\label{Preliminaries}

In this section, we provide a brief overview of the core components that underpin our proposed framework: the DreamerV3 algorithm, the theoretical basis of Kolmogorov-Arnold Networks (KANs), and the mathematical formulation of FastKAN.

\subsection{DreamerV3}

DreamerV3 is a model-based reinforcement learning algorithm that solves control tasks by interleaving three processes: data collection, world model learning, and behavior learning. The agent leverages past experiences to learn a latent dynamics model, which then serves as a simulator for training the policy via latent imagination. The overall architecture consists of several learnable components parameterized by neural networks. Broadly, the architecture can be categorized into two parts: components for world model learning and components for behavior learning.

\textbf{World Model Learning.} Following the formulation in \citet{hafner2025mastering}, the agent learns a compact latent dynamics model to represent the environment. The core is the Recurrent State-Space Model (RSSM), which maintains a deterministic recurrent state $h_t$ and a stochastic state $z_t$. The deterministic state evolves via a recurrent unit (GRU):
\begin{equation}
    h_t = f_\phi(h_{t-1}, z_{t-1}, a_{t-1})
\end{equation}

where $a_{t-1}$ denotes the previous action. To incorporate sensory information, an \textbf{encoder} maps observations $x_t$ to the stochastic posterior:
\begin{equation}
    z_t \sim q_\phi(z_t \mid h_t, x_t)
\end{equation}

In standard implementations, this encoder typically employs \textbf{CNNs} for visual inputs. 

To enable latent imagination, the model learns a \textbf{dynamics predictor} to estimate the prior state distribution $\hat{z}_t$, along with predictors for the scalar reward $\hat{r}_t$ and the episode continuation flag $\hat{c}_t$:
\begin{equation}
    \hat{z}_t \sim p_\phi(\hat{z}_t \mid h_t), \quad \hat{r}_t \sim p_\phi(\hat{r}_t \mid h_t, z_t), \quad \hat{c}_t \sim p_\phi(\hat{c}_t \mid h_t, z_t)
\end{equation}

Finally, a \textbf{decoder} reconstructs the observations to provide a learning signal:
\begin{equation}
    \hat{x}_t \sim p_\phi(\hat{x}_t \mid h_t, z_t)
\end{equation}

Crucially, the predictors ($p_\phi$) and the decoder are traditionally parametrized using \textbf{MLPs} and transposed CNNs, respectively.

\textbf{Behavior Learning.} The agent derives behaviors entirely within the latent imagination space. An \textbf{actor} network $\pi_\theta$ predicts actions to maximize returns, while a \textbf{critic} network $v_\psi$ estimates the expected value:
\begin{equation}
    a_t \sim \pi_\theta(a_t \mid s_t), \quad v_\psi(R_t \mid s_t)
\end{equation}

where $s_t = (h_t, z_t)$ represents the full latent state. Both the actor and critic are standardly instantiated as dense \textbf{MLPs}.

\subsection{Kolmogorov-Arnold Networks (KANs)}

\textbf{Architecture.} 
Kolmogorov-Arnold Networks (KANs) represent a shift from the node-centric activation of Multi-Layer Perceptrons (MLPs) to an edge-centric paradigm. Unlike MLPs, which rely on fixed activation functions on nodes and linear weights matrices $W$, KANs place \textbf{learnable univariate activation functions} on the edges and use simple summation on the nodes \citep{liu2024kan}.

Formally, for a layer with $n_{in}$ input nodes and $n_{out}$ output nodes, the activation value $x_{l+1, j}$ of the $j$-th neuron in layer $l+1$ is defined as:
\begin{equation}
    x_{l+1, j} = \sum_{i=1}^{n_{in}} \phi_{l, j, i}(x_{l, i})
\end{equation}

where $\phi_{l, j, i}$ is a learnable 1D function. This formulation eliminates standard linear weight matrices entirely.

\textbf{B-Spline Parametrization.} 
To ensure learnability, each activation function $\phi(x)$ is parametrized as a sum of a basis function $b(x)$ (typically SiLU) and a B-spline:
\begin{equation}
    \phi(x) = w_b b(x) + w_s \sum_{k} c_k B_k(x)
\end{equation}

where $c_k$ are trainable spline coefficients, and $w_b, w_s$ are scaling factors. This structure allows KANs to approximate complex functions with high parameter efficiency.

\subsection{FastKAN}

While the original KAN implementation based on B-splines offers theoretical guarantees, it suffers from significant computational inefficiencies. The recursive evaluation of B-splines is difficult to vectorize on GPUs, and the grid extension mechanism leads to a rapid increase in memory consumption. To mitigate these bottlenecks, \citet{li2024kolmogorov} proposed \textbf{FastKAN}, an efficient variant that approximates the learnable functions using Gaussian Radial Basis Functions (RBFs).

\textbf{Gaussian RBF Approximation.} 
FastKAN retains the additive structure of KANs but replaces the B-spline component with a weighted sum of Gaussian basis functions. For a learnable activation function $\phi(x)$, the formulation becomes:
\begin{equation}
    \phi(x) = w_b b(x) + w_s \sum_{k=1}^{G} c_k \exp\left(-\left(\frac{x - \mu_k}{h}\right)^2\right)
\end{equation}

where $G$ is the number of Gaussian basis functions. We explicitly denote this hyperparameter as $G$ (Grid Size) rather than the conventional $C$ (Centers) to maintain terminological consistency with the original spline-based KANs and the official FastKAN implementation (where it is defined as \texttt{num\_grids}). Furthermore, $\mu_k$ represent the fixed grid centers uniformly distributed over the input range, and $h$ is a hyperparameter controlling the bandwidth of the Gaussian functions. Similar to the original KAN, $c_k$ represents the learnable coefficients, while $w_b$ and $w_s$ are scaling parameters.

\textbf{Computational Efficiency.} 
The key advantage of this formulation is that the RBF evaluation can be vectorized as standard matrix multiplications, which are highly optimized in modern deep learning frameworks (e.g., PyTorch/CUDA). Unlike B-splines, which require conditional control flows for grid intervals, the RBF layer processes inputs in parallel, offering substantial improvements in training throughput and memory efficiency.

\section{Method}
\label{Method}

\subsection{Overall Architecture}

We introduce \textbf{KAN-Dreamer}, a hybrid architecture that integrates the distinct function approximation capabilities of Kolmogorov-Arnold Networks into the model-based reinforcement learning framework. While preserving the macroscopic data flow and training objectives of DreamerV3, we adopt a modular backbone strategy. This design conceptually decouples the architecture into two functional categories: \textbf{recurrent dynamics modeling} (which remains fixed) and \textbf{feed-forward function approximation} (the target of our investigation). By isolating these roles, KAN-Dreamer allows for the flexible instantiation of non-recurrent sub-networks using either KANs, FastKANs, or standard MLPs and CNNs. This flexibility enables a systematic benchmarking of KANs' efficacy within a complex RL system, without altering the fundamental algorithmic logic. The specific architectural configurations for the World Model and Behavior Learning components are detailed in the following subsections.

\subsection{World Model Architecture}
\label{Method:World Model Architecture}

\begin{figure}[t]
  \centering
  \includegraphics[width=0.9\textwidth]{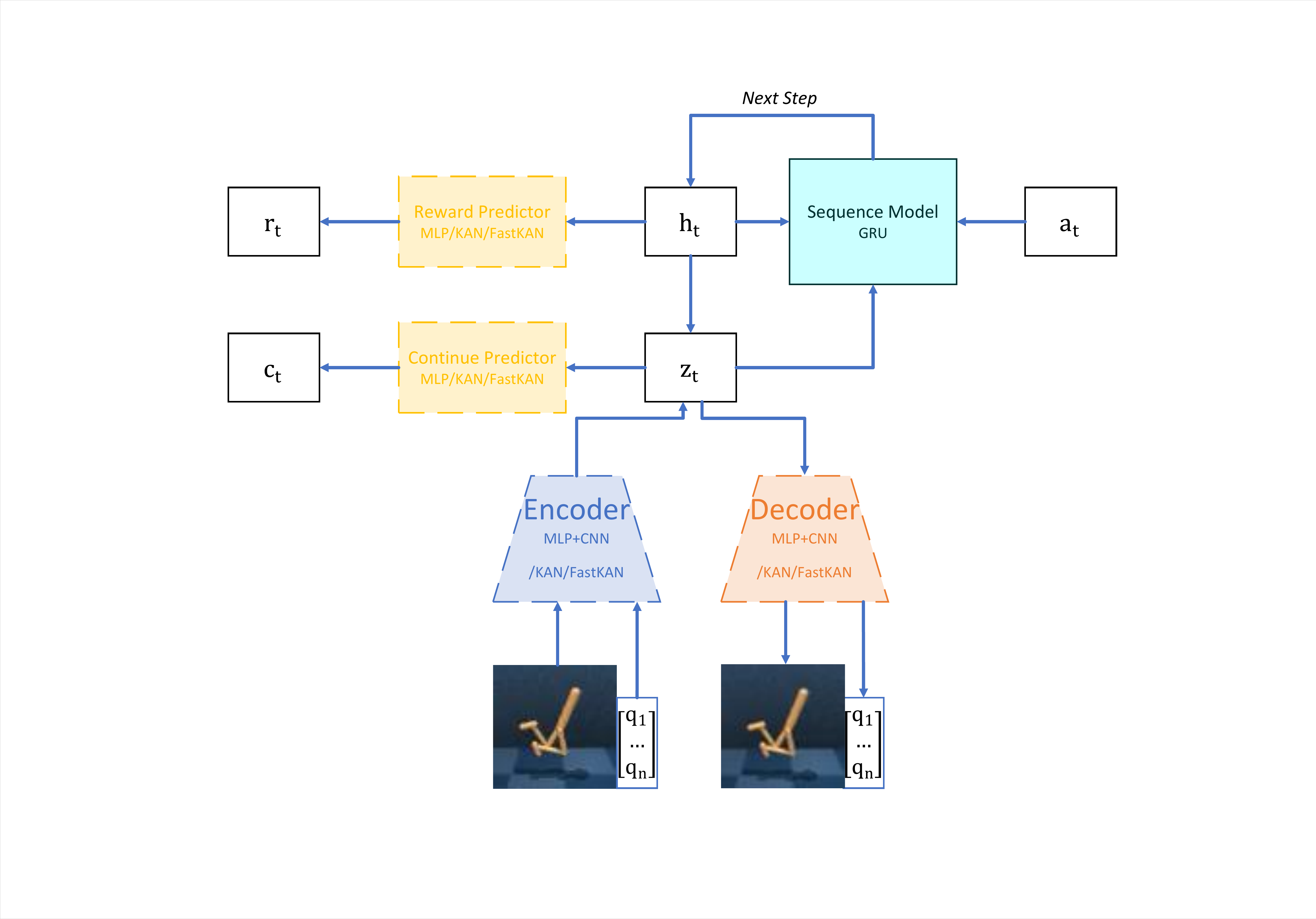}
  \caption{\textbf{Architecture of the KAN-Dreamer World Model.} 
  (a) \textbf{Variables:} White boxes denote data tensors: $z_t$ (stochastic latent state), $h_t$ (deterministic recurrent state), $r_t$ (reward), $c_t$ (continuation flag), and $a_t$ (action). 
  (b) \textbf{Components:} Colored boxes represent learnable modules. \textbf{Dashed borders} indicate configurable modules (Encoder, Decoder, Predictors) where the backbone can be instantiated as \textbf{standard architectures (MLP/CNN)}, KAN, or FastKAN. The \textbf{solid-bordered} Sequence Model (cyan) retains the fixed GRU architecture for stability. Blue arrows indicate the data flow during the forward pass. \textbf{Note: Proprioceptive inputs $q$ are depicted for completeness but were masked out in our visual-only experiments.}}
  \label{fig:world_model} 
\end{figure}

The basic structure of the world model is illustrated in Figure~\ref{fig:world_model}. Specifically, we redesign the World Model's Encoder, Decoder, Reward Predictor, and Continue Predictor to support interchangeable backbones. This flexible design allows the \textbf{vector-based components} (Reward and Continue Predictors) to be instantiated using either standard \textbf{MLPs}, KANs, or FastKANs, while enabling the \textbf{visual components} (Encoder, Decoder) to switch between standard \textbf{CNNs}, KANs, and FastKANs.

\textbf{Visual Processing Adaptation.} 
A notable structural modification is applied to the visual Encoder and Decoder. Drawing inspiration from the \textbf{Kolmogorov-Arnold Auto-Encoder (KAE)} \citep{yu2024kae}, we replace the standard Convolutional Neural Networks (CNNs) with fully connected KAN layers. 
However, distinct from the polynomial basis functions originally proposed in KAE, we prioritize \textbf{B-splines (KAN)} and \textbf{Gaussian RBFs (FastKAN)}. This choice is motivated by the superior local flexibility and numerical stability of splines and RBFs compared to global polynomials, which are prone to oscillation in deep architectures.
The processing logic is adapted as follows: for encoding, high-dimensional image observations are \textbf{flattened} into 1D vectors before being fed into the KAN backbone; for decoding, the latent features are projected by the KAN layers into a flat vector and subsequently \textbf{reshaped} to match the original image dimensions for pixel-level reconstruction.

\textbf{Latent Prediction Adaptation.}
In contrast to the visual components, the Reward and Continue predictors operate on low-dimensional latent states ($h_t, z_t$). Consequently, these modules require no structural reshaping. We implement them as direct \textbf{drop-in replacements}, where the standard MLP layers are substituted with KAN or FastKAN layers to map the latent state to scalar rewards and continuation probabilities.

Crucially, we deliberately retain the original GRU-based Sequence Model (RSSM). This component serves as the core engine for latent imagination, imposing stringent demands on both prediction fidelity and computational throughput to ensure stable and efficient policy learning. We adhere to the GRU baseline because the efficiency and efficacy of KANs in recurrent settings remain under-explored, and no established KAN-based architecture currently exists to effectively replace the GRU without compromising training stability.

\subsection{Actor-Critic Architecture}

\begin{figure}[h]
  \centering
  \includegraphics[width=0.6\textwidth]{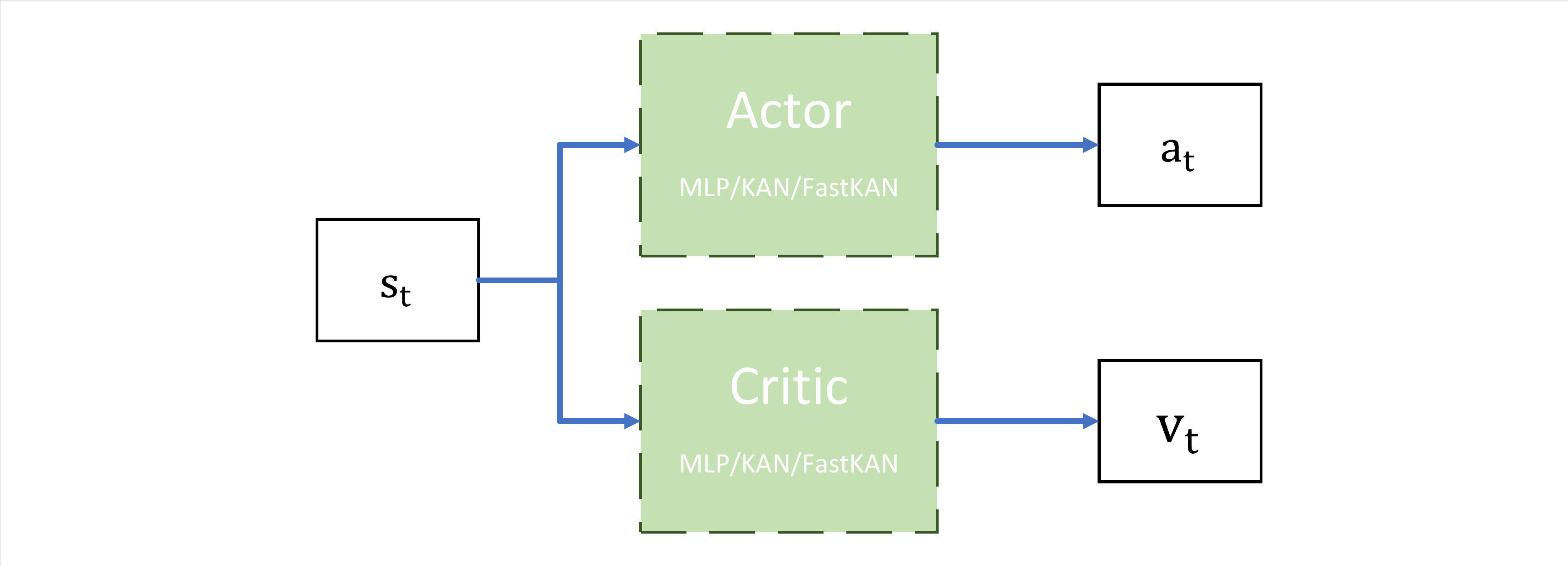}
  \caption{Actor-Critic architecture}
  \label{fig:actor-critic} 
\end{figure}

The basic structure of the Actor-Critic architecture is illustrated in Figure~\ref{fig:actor-critic}.In the standard DreamerV3 formulation, both the Actor network ($\pi_\theta$) and the Critic network ($v_\psi$) are parametrized using dense \textbf{Multi-Layer Perceptrons (MLPs)}.Given that these components, analogous to the Reward and Continue predictors discussed in Section \ref{Method:World Model Architecture}, operate exclusively on the low-dimensional latent state $s_t = (h_t, z_t)$, they require no spatial transformations. Consequently, we adopt an identical integration strategy: we implement KAN and FastKAN layers as direct \textbf{drop-in replacements} for the MLP backbones. This configuration allows us to isolate and evaluate whether the learnable univariate functions of KANs can approximate complex policy boundaries and value landscapes more efficiently than traditional static activations within the imagination loop.

\subsection{Implementation and Architectural Adaptation}

Implementing KAN-Dreamer presents a practical challenge due to the framework disparity: the original DreamerV3 is implemented in JAX \citep{hafner2025mastering}, whereas current KAN and FastKAN libraries are predominantly PyTorch-native, making direct integration infeasible. To bridge this gap and ensure efficient computation within the World Model, we propose a tailored implementation of KAN and FastKAN, introducing specific structural modifications to align with the JAX-based DreamerV3 architecture.

Furthermore, given the distinct operational objectives—shifting from symbolic regression to high-dimensional latent dynamics modeling—simply inheriting the default hyperparameters (specifically the grid size $G$ and spline order $k$) proves suboptimal. Consequently, we recalibrate these parameters to balance expressivity with the computational constraints of the World Model.

\textbf{Tensorized Computation via JAX.} 
Departing from original implementations that often process input dimensions sequentially, our adaptation leverages JAX's powerful vectorization capabilities. For \textbf{KAN}, we pack all spline coefficients into a monolithic 3D tensor and perform the B-spline aggregation using a single \texttt{einsum} operation. Similarly, for \textbf{FastKAN}, we maintain the Radial Basis Function (RBF) outputs in their native tensor structure $[N, D, G]$ (where $G$ denotes the number of basis functions) and apply weights via tensor contraction. This \textbf{fully vectorized approach} minimizes memory overhead and maximizes GPU utilization, contributing to the high training throughput observed in our experiments.

\textbf{Simplified Grid Management and Initialization.} 
To maintain training stability within the complex latent dynamics of DreamerV3, we simplified the adaptive mechanisms of the original KAN. Crucially, we employ \textbf{fixed uniform grids} for spline interpolation rather than the data-dependent adaptive grid updates used in the original paper, which avoids non-stationary gradients during online RL. Specifically, we configure the \textbf{KAN} layers with a grid size of $G=8$ and a spline order of $k=3$, defined over a clamped input range of $[-5.0, 5.0]$. For \textbf{FastKAN}, we deploy $G=8$ Gaussian centers (aligning with the grid size terminology) distributed over $[-2.0, 2.0]$. In this formulation, the spline order concept is effectively substituted by the RBF bandwidth, which is automatically derived from the uniform grid interval. While this fixed-grid approach may theoretically limit the approximation precision compared to adaptive grids, it significantly stabilizes the gradient flow. To further ensure stability, we paid close attention to initialization. For the linear transformations within FastKAN (the base branch), we utilize the native linear layers from the Dreamer codebase (\texttt{embodied.jax.nets.Linear}) for consistent weight and bias handling. For the activation function scaling parameters, we adopted distinct strategies: in our \textbf{KAN} implementation, the base weight ($w_b$) is initialized from a normal distribution with a standard deviation of approximately $0.5/\sqrt{D_{in}}$, while the spline weight ($w_s$) is fixed at a value scaled by $1/\sqrt{D_{in}}$. In our \textbf{FastKAN} implementation, both scaling weights ($w_b$ and $w_s$) are kept fixed at 1.0, relying on the standard initialization of the linear layers. Auxiliary features such as masking and pruning were omitted to prioritize runtime performance.

\textbf{A Note on Terminology.} The implementations described above—featuring JAX-based vectorization, fixed grids, and simplified initialization—represent specific adaptations of the original architectures tailored for our experimental context. For clarity and brevity, we will refer to these modified versions simply as \textbf{"KAN"} and \textbf{"FastKAN"} throughout the rest of this paper.

\section{Experiments}
\label{Experiments}

\subsection{Experimental Setup}
\label{Experiments:Experimental Setup}

\textbf{Task and Environment.} 
We benchmark our proposed architecture on the \texttt{walker\_walk} task from the DeepMind Control Suite (DMC) \citep{tassa2018deepmind}. Adhering to the "learning from pixels" philosophy of DreamerV3, we configure the environment to provide \textbf{pure image observations}, strictly excluding any proprioceptive state vectors (e.g., joint angles). This setting rigorously tests the capability of KAN-based components in processing high-dimensional visual data.

\textbf{Ablation Design.} 
To isolate the impact of KANs and FastKANs on specific subsystems, we conduct a systematic component-wise ablation study divided into three groups:
\begin{itemize}
    \item \textbf{Perception Group:} Comparing the Baseline against variants where the Visual Encoder and Decoder are replaced by KANs or FastKANs.
    \item \textbf{Prediction Group:} Comparing the Baseline against variants with KAN-based or FastKAN-based Reward and Continue predictors.
    \item \textbf{Behavior Group:} Comparing the Baseline against variants where the Actor and Critic networks are instantiated as KANs or FastKANs.
\end{itemize}

\textbf{Model Configuration.} 
To ensure a fair comparison, we adopt an \textbf{iso-parameter} evaluation protocol. We adjust the hidden dimensions (and grid parameters where applicable) of all variants to maintain a total parameter count of approximately \textbf{10.5M ($\pm$0.1M)}. Detailed architectural specifications (e.g., \textbf{CNN depth, layer units, and grid size}) are provided in \textbf{Appendix A}. All other training hyperparameters (e.g., learning rate, batch size, horizon) strictly follow the official DreamerV3 configuration to ensure reproducibility, as detailed in \textbf{Appendix B}.

\textbf{Evaluation Metrics.} 
We structure our experimental analysis into three complementary dimensions:
\begin{itemize}
    \item \textbf{Quantitative Analysis:} We focus on two key indicators: (1) \textbf{Asymptotic Performance}, measured by the final cumulative return after \textbf{1.1M environment steps}, aligning with the standard benchmarking protocol of DreamerV3. A score threshold of $\ge 900$ is established to validate the effectiveness of the architectural modifications. (2) \textbf{Computational Efficiency}, assessed by two distinct throughput metrics. \textit{FPS/Train} measures the gradient update speed, isolating the computational overhead of the KAN/MLP backbones. \textit{FPS/Policy} measures the overall training throughput (including environment interaction), which serves as a proxy for the total wall-clock training time.
    \item \textbf{Training Dynamics and Stability:} Beyond point estimates, we provide a multi-faceted analysis of the learning process: (1) \textbf{Sample Efficiency}, visualized via cumulative reward curves against environment steps; (2) \textbf{Wall-Clock Efficiency}, evaluating the trade-off between sample complexity and physical runtime; and (3) \textbf{Component Optimization}, where we examine the training loss trajectories (e.g., reconstruction loss, prediction error) to diagnose convergence behaviors.
\end{itemize}

\subsection{Implementation Details}

\textbf{Computational Resources.} 
All experiments were conducted on a server equipped with an \textbf{NVIDIA A800 Tensor Core GPU} and an AMD EPYC 9654 CPU (7 cores allocated per run).

\textbf{Software Stack.} 
The implementation runs on Anolis OS 8.6 with Python 3.11. Our code is built upon the \textbf{JAX framework} (version 0.4.33, CUDA 12 support), ensuring compatibility with the official DreamerV3 codebase.

\subsection{Quantitative Analysis}

\begin{table}[t]
  \centering
  \caption{\textbf{Component-wise Ablation Study.} 
  We evaluate the impact of KAN/FastKAN on specific subsystems. 
  Symbols: $\circ$ = Original (MLP/CNN), $\bullet$ = \textbf{KAN}, $\blacksquare$ = \textbf{FastKAN}.
  \textbf{FPS (Pol.)}: Environment steps per second.
  \textbf{FPS (Trn.)}: Training throughput.
  \textit{Note: All model variants are controlled to have approximately 10.5M parameters.}}
  \label{tab:component_ablation}
  
  \begin{subtable}{\textwidth}
    \centering
    \caption{\textbf{Impact on Perception (Visual Encoder \& Decoder)}}
    \label{tab:ablation_vis}
    \begin{tabular}{l c | c | ccc}
      \toprule
      & \textbf{Visual} & \textbf{Perf.} & \multicolumn{3}{c}{\textbf{Efficiency}} \\
      \textbf{Model ID} & \textbf{Enc./Dec.} & \textbf{Final Score} & \textbf{FPS (Pol.)} & \textbf{FPS (Trn.)} & \textbf{Params} \\
      \midrule
      Baseline & $\circ$ & 977 & 62 & 15.9k & 10.49M \\
      \midrule
      KAN-Vis. & $\bullet$ & 940 & 72 & 18.5k & 10.40M \\
      \midrule
      FKAN-Vis. & $\blacksquare$ & 950 & 89 & 22.8k & 10.51M \\
      \bottomrule
    \end{tabular}
  \end{subtable}

  \vspace{1em} 

  \begin{subtable}{\textwidth}
    \centering
    \caption{\textbf{Impact on Prediction Heads (Reward \& Continue)}}
    \label{tab:ablation_pred}
    \begin{tabular}{l c | c | ccc}
      \toprule
      & \textbf{Reward/} & \textbf{Perf.} & \multicolumn{3}{c}{\textbf{Efficiency}} \\
      \textbf{Model ID} & \textbf{Cont. Pred.} & \textbf{Final Score} & \textbf{FPS (Pol.)} & \textbf{FPS (Trn.)} & \textbf{Params} \\
      \midrule
      Baseline & $\circ$ & 977 & 62 & 15.9k & 10.49M \\
      \midrule
      KAN-Pred & $\bullet$ & 965 & 45 & 11.5k & 10.45M \\
      FKAN-Pred & $\blacksquare$ & 951 & 61 & 15.5k & 10.51M \\
      \bottomrule
    \end{tabular}
  \end{subtable}

  \vspace{1em} 

  \begin{subtable}{\textwidth}
    \centering
    \caption{\textbf{Impact on Behavior Learning (Actor \& Critic)}}
    \label{tab:ablation_ac}
    \begin{tabular}{l c | c | ccc}
      \toprule
      & \textbf{Actor/} & \textbf{Perf.} & \multicolumn{3}{c}{\textbf{Efficiency}} \\
      \textbf{Model ID} & \textbf{Critic} & \textbf{Final Score} & \textbf{FPS (Pol.)} & \textbf{FPS (Trn.)} & \textbf{Params} \\
      \midrule
      Baseline & $\circ$ & 977 & 62 & 15.9k & 10.49M \\
      \midrule
      KAN-AC & $\bullet$ & 957 & 31 & 7.8k & 10.48M \\
      FKAN-AC & $\blacksquare$ & 964 & 57 & 14.5k & 10.47M \\
      \bottomrule
    \end{tabular}
  \end{subtable}
\end{table}

\textbf{Baseline Performance.} 
We first establish the performance benchmark using the standard DreamerV3 algorithm. As shown in the first row of Table \ref{tab:component_ablation}(a-c), under the iso-parameter setting ($\sim$10.5M), the baseline achieves a final score of \textbf{977}, with an interaction speed of \textbf{62 FPS (Pol.)} and a training throughput of \textbf{15.9k FPS (Trn.)}. These metrics serve as the reference for the subsequent comparisons.

\textbf{Visual Perception.} 
Table \ref{tab:ablation_vis} presents the results for the Perception Group. Replacing the CNN-based encoder and decoder with KAN or FastKAN layers (using flattened inputs) yields final scores of 940 and 950, respectively. Regarding computational efficiency, both variants exhibit higher throughput metrics than the CNN baseline. Specifically, FastKAN achieves the highest speeds (89 FPS Pol. / 22.8k FPS Trn.), followed by KAN and the Baseline (\textbf{FastKAN > KAN > Baseline}).

\textbf{Latent Prediction.} 
As shown in Table \ref{tab:ablation_pred}, substituting the MLP-based Reward and Continue predictors with KANs or FastKANs results in asymptotic returns exceeding 950 for both variants. In terms of efficiency, the standard KAN implementation shows a reduction in interaction speed to 45 FPS. In contrast, the FastKAN variant records a speed of 61 FPS, which is comparable to the baseline's 62 FPS, following the trend \textbf{Baseline $\approx$ FastKAN > KAN}.

\textbf{Policy Optimization.} 
Table \ref{tab:ablation_ac} reports the results for the Behavior Learning components. Both KAN-AC and FKAN-AC variants reach final scores of 957 and 964, respectively. The computational metrics show a distinct difference between the two variants: the standard KAN model operates at 31 FPS (interaction) and 7.8k FPS (training), whereas the FastKAN model maintains speeds of 57 FPS and 14.5k FPS, remaining close to the MLP baseline figures (\textbf{Baseline > FastKAN > KAN}).

\subsection{Training Dynamics}

\subsubsection{Sample Efficiency and Convergence}
\label{Sample Efficiency and Convergence}

\begin{figure}[t]
    \centering
    
    \begin{subfigure}{\textwidth}
        \centering
        \includegraphics[width=\textwidth, trim=10 280 2 280, clip]{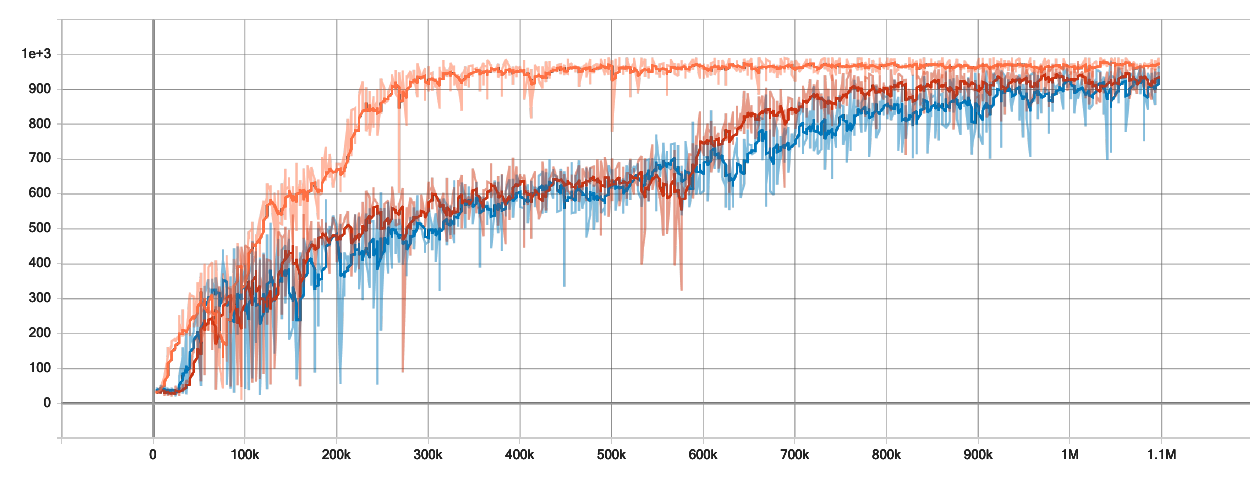}
        \caption{\textbf{Perception:} Comparison of Baseline (orange), KAN-Vis (blue), and FKAN-Vis (red).}
        \label{fig:vis_score}
    \end{subfigure}
    
    \vspace{0.5em}

    \begin{subfigure}{\textwidth}
        \centering
        \includegraphics[width=\textwidth, trim=10 280 2 280, clip]{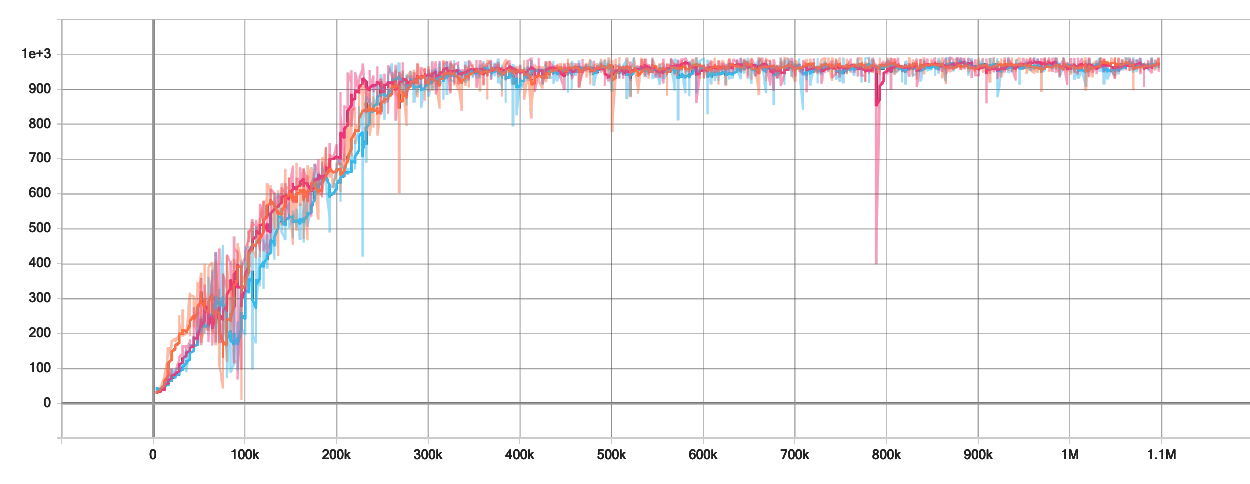}
        \caption{\textbf{Prediction:} Comparison of Baseline (orange), KAN-Pred (light blue), and FKAN-Pred (pink).}
        \label{fig:pred_score}
    \end{subfigure}
    
    \vspace{0.5em}

    \begin{subfigure}{\textwidth}
        \centering
        \includegraphics[width=\textwidth, trim=10 280 2 280, clip]{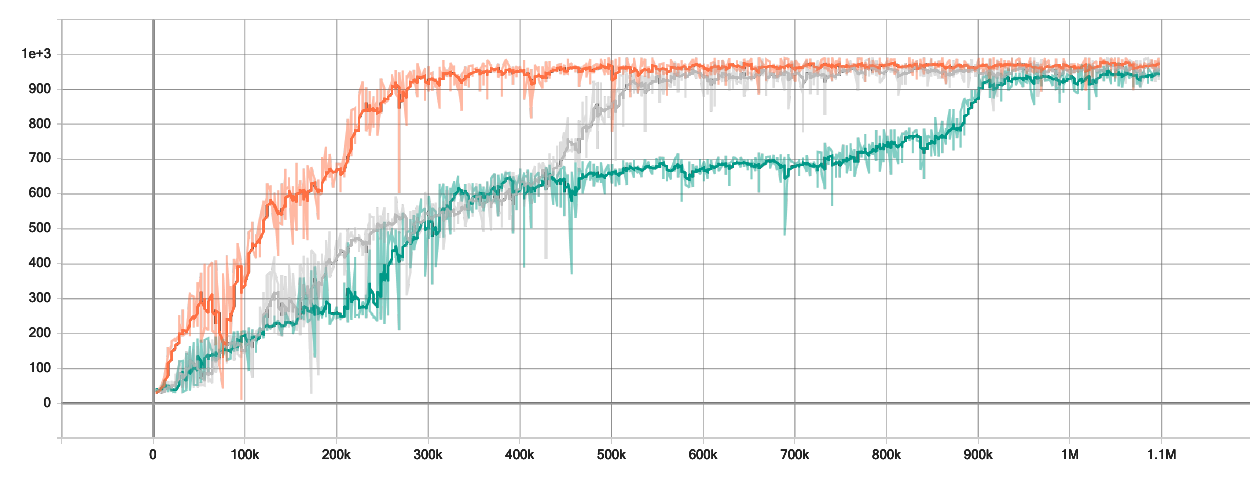}
        \caption{\textbf{Behavior:} Comparison of Baseline (orange), KAN-AC (green), and FKAN-AC (gray).}
        \label{fig:ac_score}
    \end{subfigure}

    \caption{\textbf{Training Dynamics across Component Groups.} 
    Darker lines indicate smoothed curves (factor 0.8), while lighter traces show raw data.}
    \label{fig:training_dynamics_all}
\end{figure}

We analyze the learning trajectories of cumulative rewards to understand the sample efficiency and stability of different architectural variants. \textbf{Visualization Setup:} To clearly visualize the overall convergence trends amidst training stochasticity, all reward curves are applied with a smoothing factor of 0.8. As shown in Figure \ref{fig:training_dynamics_all}, the \textbf{solid (darker) lines} represent the smoothed trajectories, while the \textbf{translucent (lighter) traces} in the background depict the raw, unsmoothed data points.

\textbf{Baseline Reference.}
First, we examine the \textbf{Baseline} (orange curves) to establish a performance standard. Consistent with DreamerV3's state-of-the-art capabilities, the baseline demonstrates rapid convergence, typically crossing the high-performance threshold (reward $\ge 900$) within approximately \textbf{250k} steps. Moreover, its raw reward traces remain tightly clustered around the mean, indicating a highly stable learning process.

\textbf{Visual Perception.} 
As illustrated in Figure \ref{fig:vis_score}, the smoothed trend lines show a distinct difference in sample efficiency between the KAN-based variants and the baseline. Both variants require a substantially higher number of environment steps to exceed a return of 900 (approximately \textbf{930k} steps for KAN and \textbf{750k} steps for FastKAN). Furthermore, the raw data traces reveal that KAN-based encoders exhibit larger reward fluctuations throughout the training process compared to the CNN baseline, reflecting higher variance in the learned visual representation.

\textbf{Latent Prediction.} 
In contrast to perception, the learning curves for the Prediction Group (Figure \ref{fig:pred_score}) demonstrate that KAN-based predictors achieve convergence profiles comparable to the MLP baseline. All models converge around \textbf{250k} steps. Notably, the FastKAN variant exhibits a slight advantage, crossing the 900-reward threshold as early as \textbf{220k} steps. Regarding stability, the raw data traces show that the variance of KAN and FastKAN variants remains within a range similar to that of the baseline, showing that low-dimensional regression heads can be replaced without affecting training stability.

\textbf{Policy Optimization.} 
Figure \ref{fig:ac_score} highlights a gap in convergence speed for the Behavior Group. Both KAN and FastKAN variants converge slower than the baseline. Specifically, FastKAN reaches the asymptotic performance threshold at approximately \textbf{520k} steps, while the standard KAN requires nearly \textbf{900k} steps. Despite this delay in convergence, the training stability—as evidenced by the raw reward fluctuations—remains controlled and comparable to the Latent Prediction experiments, avoiding the high variance observed in the Visual Perception task.

\subsubsection{Wall-Clock Efficiency Analysis}

\begin{figure}[t]
    \centering
    
    \begin{subfigure}{\textwidth}
        \centering
        \includegraphics[width=\textwidth, trim=10 280 2 280, clip]{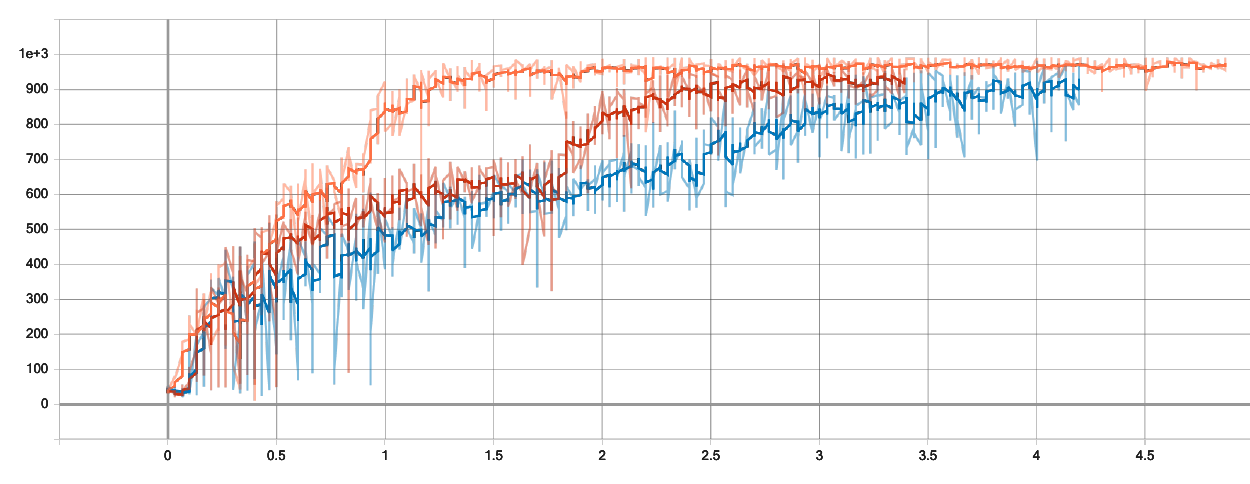}
        \caption{\textbf{Perception:} Comparison of Baseline (orange), KAN-Vis (blue), and FKAN-Vis (red).}
        \label{fig:vis_score_time}
    \end{subfigure}
    
    \vspace{0.5em}

    \begin{subfigure}{\textwidth}
        \centering
        \includegraphics[width=\textwidth, trim=10 280 2 280, clip]{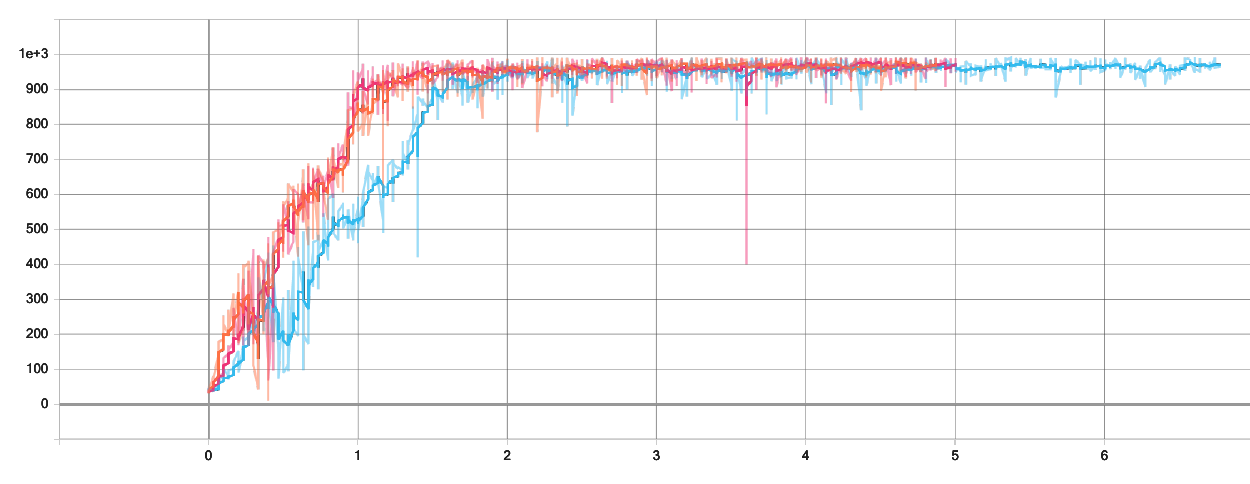}
        \caption{\textbf{Prediction:} Comparison of Baseline (orange), KAN-Pred (light blue), and FKAN-Pred (pink).}
        \label{fig:pred_score_time}
    \end{subfigure}
    
    \vspace{0.5em}

    \begin{subfigure}{\textwidth}
        \centering
        \includegraphics[width=\textwidth, trim=10 280 2 280, clip]{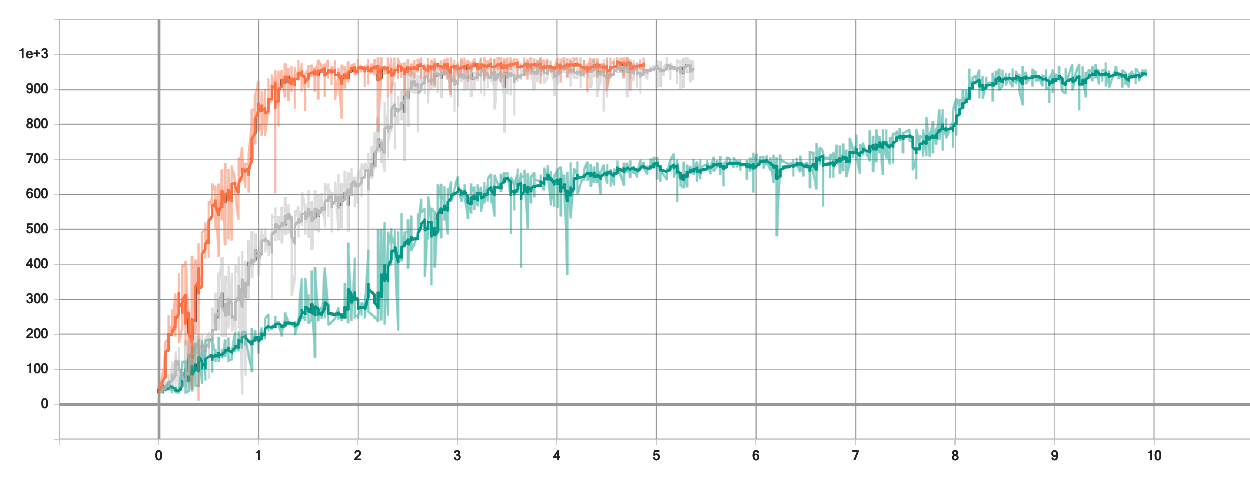}
        \caption{\textbf{Behavior:} Comparison of Baseline (orange), KAN-AC (green), and FKAN-AC (gray).}
        \label{fig:ac_score_time}
    \end{subfigure}

    \caption{\textbf{Wall-Clock Training Efficiency.} 
    Darker lines indicate smoothed curves (factor 0.8), while lighter traces show raw data.}
    \label{fig:time_efficiency_all}
\end{figure}

While sample efficiency focuses on data economy, the total physical training duration is \textbf{equally critical for rapid experimental iteration and practical scalability}. We plot the cumulative returns against wall-clock time (in hours) to evaluate this efficiency-performance trade-off. \textbf{Visualization Setup:} Similar to Section \ref{Sample Efficiency and Convergence}, the curves are smoothed with a factor of 0.8. We focus our analysis here exclusively on the \textbf{time-to-convergence profile} shown in Figure \ref{fig:time_efficiency_all}.

\textbf{Baseline Reference.}
The standard DreamerV3 (orange curves) demonstrates high time efficiency across all tasks. As a reference point, the baseline consistently reaches the high-performance threshold (reward $\ge 900$) within approximately \textbf{1.1 hours} of physical runtime.

\textbf{Visual Perception.} 
As shown in Figure \ref{fig:vis_score_time}, both KAN-based variants require longer physical time to converge compared to the baseline. Specifically, the FastKAN variant reaches the performance threshold in about \textbf{2.5 hours}, while the standard KAN variant takes approximately \textbf{3.5 hours}.

\textbf{Latent Prediction.} 
Figure \ref{fig:pred_score_time} illustrates the efficiency in the Prediction Group. The standard KAN variant exhibits a slight delay, requiring approximately \textbf{1.5 hours} to exceed a return of 900. In contrast, the FastKAN variant matches the baseline's efficiency, reaching the high-performance zone in just under \textbf{1 hour}, slightly faster than the 1.1-hour benchmark of the baseline.

\textbf{Policy Optimization.} 
In the Behavior Group (Figure \ref{fig:ac_score_time}), both KAN-based architectures show a marked increase in time-to-convergence. The standard KAN implementation requires over \textbf{8 hours} of training to approach convergence. The FastKAN variant improves upon the standard KAN but still trails the baseline, requiring approximately \textbf{2.5 hours} to reach the 900-reward threshold.

\subsubsection{Component Optimization Dynamics}

To investigate the training behavior of the replaced components, we examine the training loss curves.

\begin{figure}[H] 
    \centering
    \includegraphics[width=\textwidth, trim=10 280 2 280, clip]{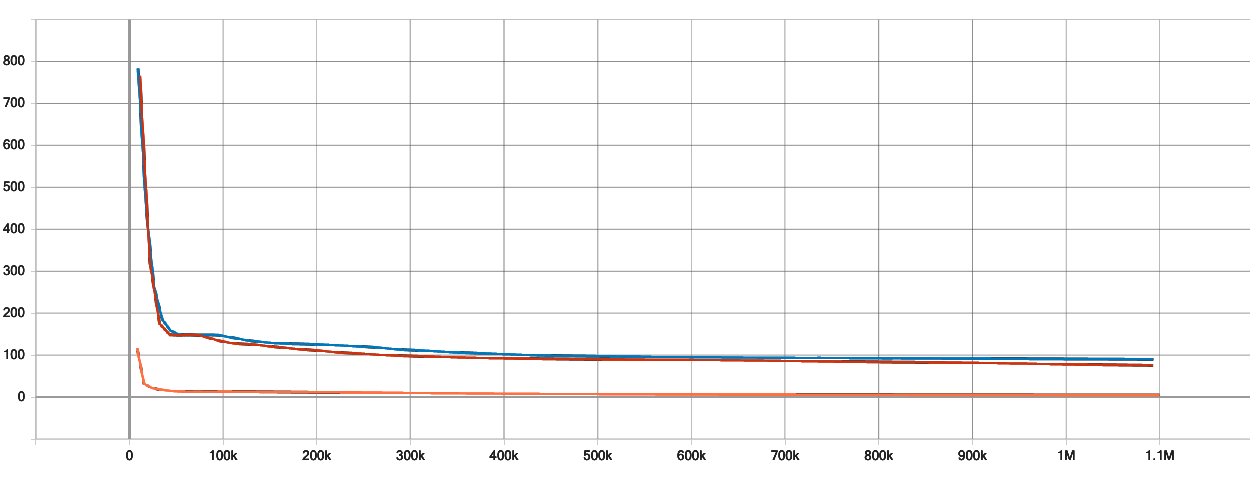}
    \caption{\textbf{Image Reconstruction Loss.} 
    Comparison of Baseline (orange), KAN-Vis (blue), and FKAN-Vis (red).}
    \label{fig:loss_image}
\end{figure}

\textbf{Visual Perception.} 
Since the primary objective of the autoencoder is to compress and reconstruct visual observations, we track the \textbf{image reconstruction loss} to assess representation quality. As illustrated in Figure \ref{fig:loss_image}, the baseline algorithm achieves near-perfect convergence, with the loss approaching zero. In stark contrast, both KAN and FastKAN variants plateau at a significantly higher level (approximately 100) and fail to decrease further, indicating a persistent gap in reconstruction fidelity.

\begin{figure}[H]
    \centering
    \begin{subfigure}{0.49\textwidth}
        \centering
        \includegraphics[width=\textwidth, trim=10 200 2 200, clip]{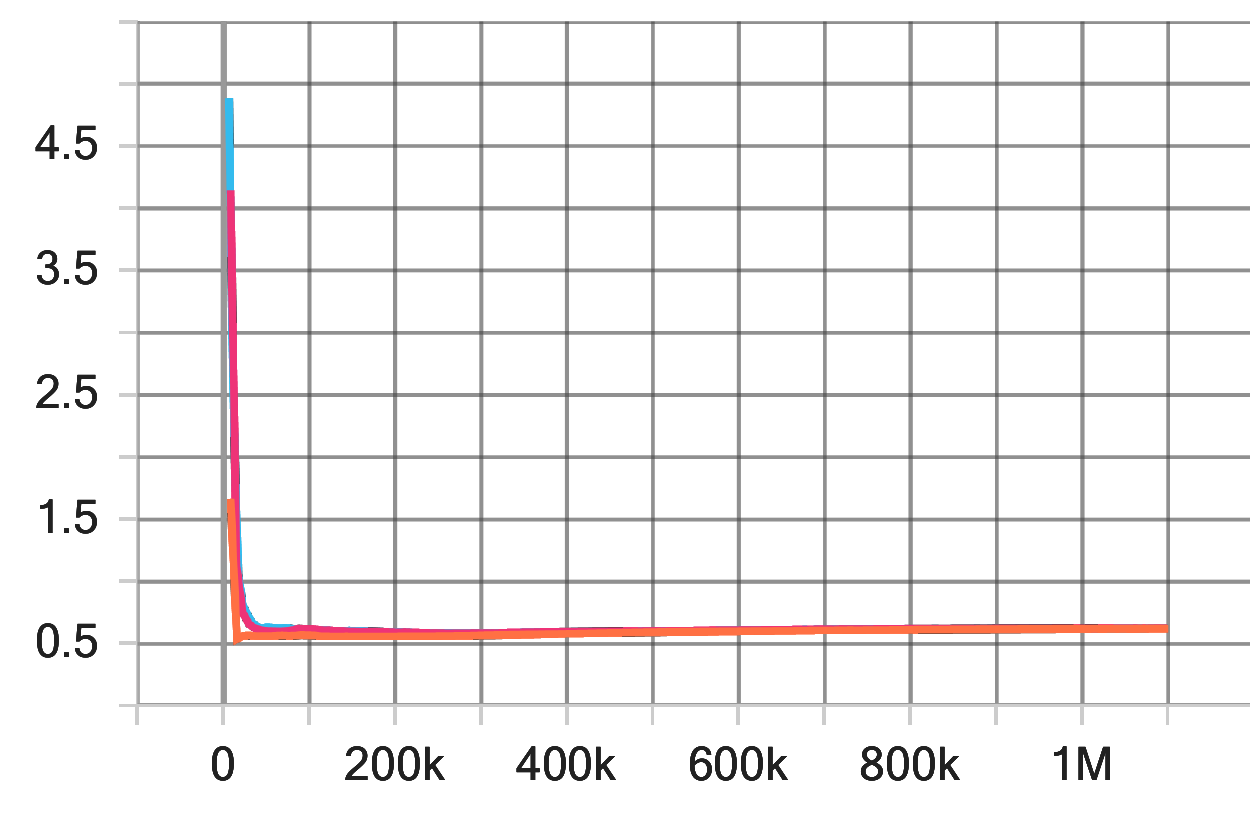}
        \caption{Reward Prediction Loss}
        \label{fig:loss_rew}
    \end{subfigure}
    \hfill
    \begin{subfigure}{0.49\textwidth}
        \centering
        \includegraphics[width=\textwidth, trim=10 200 2 200, clip]{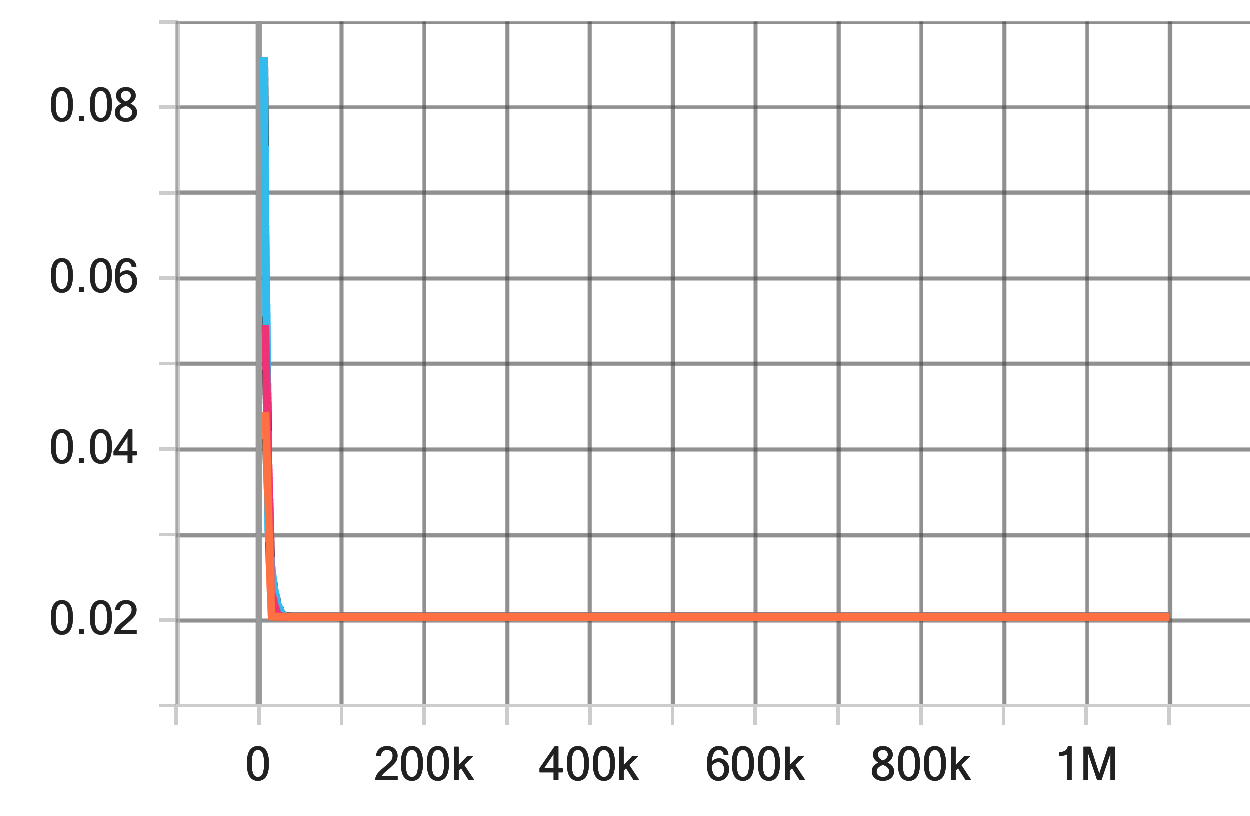}
        \caption{Continue Prediction Loss}
        \label{fig:loss_con}
    \end{subfigure}
    \caption{\textbf{Latent Prediction Losses.} 
    Comparison of Baseline (orange), KAN-Pred (light blue), and FKAN-Pred (pink).}
    \label{fig:loss_prediction}
\end{figure}

\textbf{Latent Prediction.} 
The Reward and Continue predictors aim to estimate scalar rewards and termination flags, respectively. Their optimization progress is shown in Figure \ref{fig:loss_prediction}. As seen in Figures \ref{fig:loss_rew} and \ref{fig:loss_con}, all three models ultimately converge to similar low loss values (Reward $\approx 0.625$, Continue $\approx 0.020$). However, the KAN and FastKAN variants exhibit a slightly slower initial convergence rate compared to the MLP baseline before reaching parity.

\begin{figure}[H]
    \centering
    \begin{subfigure}{0.49\textwidth}
        \centering
        \includegraphics[width=\textwidth, trim=10 200 2 200, clip]{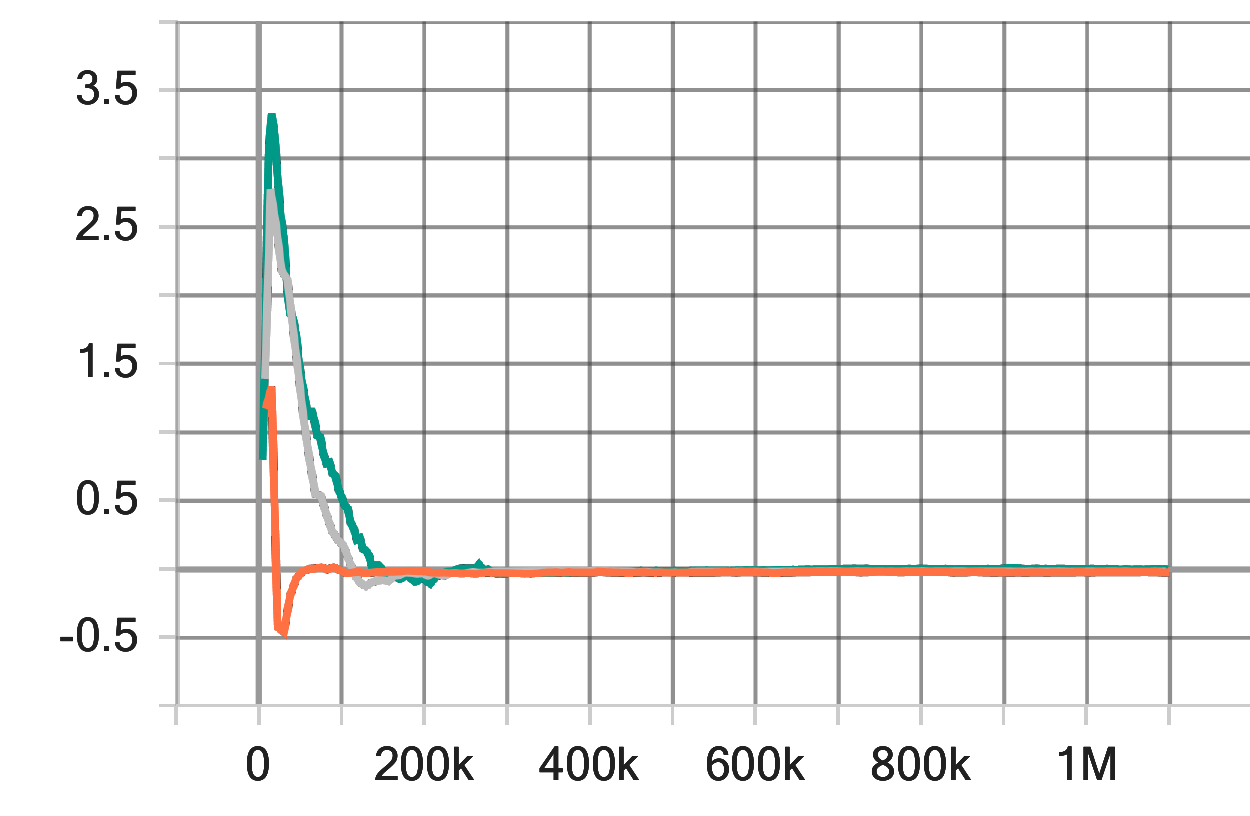}
        \caption{Actor Policy Loss}
        \label{fig:loss_policy}
    \end{subfigure}
    \hfill
    \begin{subfigure}{0.49\textwidth}
        \centering
        \includegraphics[width=\textwidth, trim=10 200 2 200, clip]{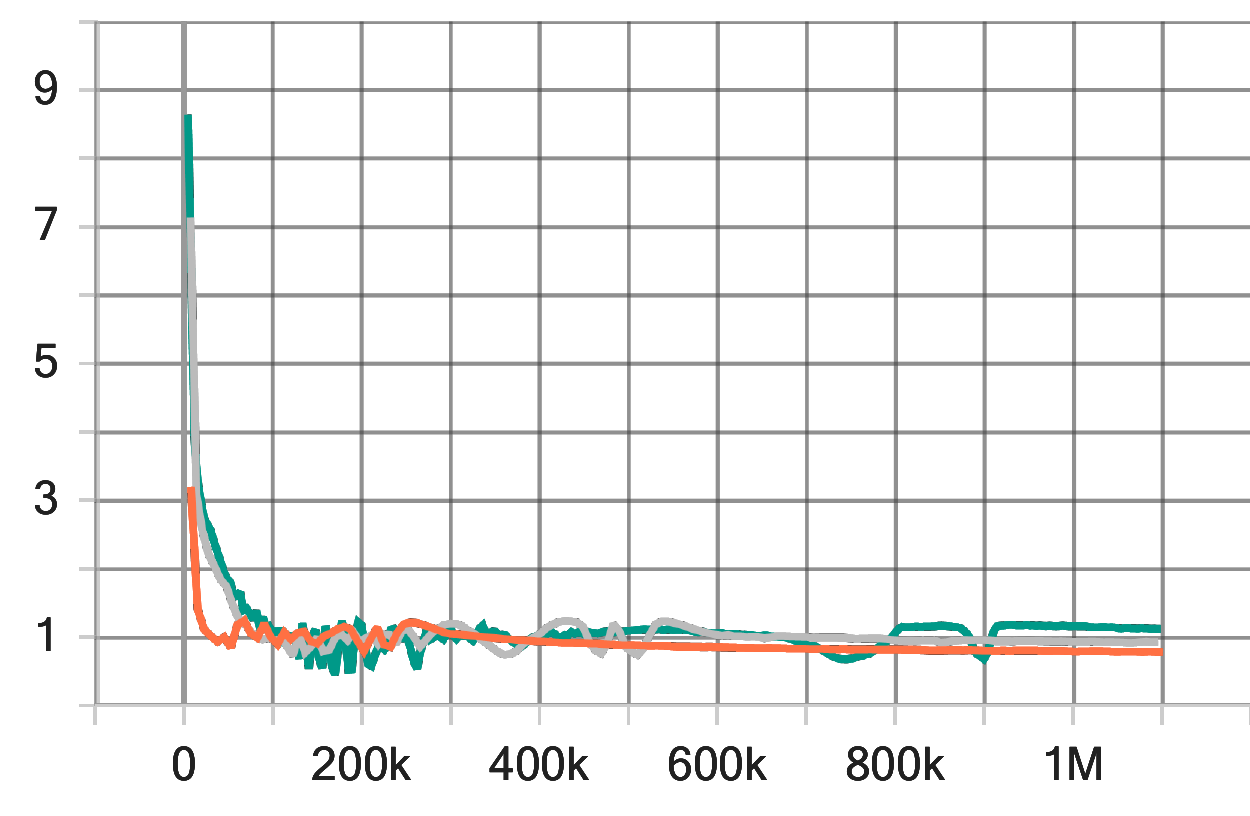}
        \caption{Critic Value Loss}
        \label{fig:loss_value}
    \end{subfigure}
    \caption{\textbf{Behavior Learning Losses.} 
    Comparison of Baseline (orange), KAN-AC (green), and FKAN-AC (gray).}
    \label{fig:loss_behavior}
\end{figure}

\textbf{Policy Optimization.} 
We analyze the \textbf{policy loss} (for the Actor) and \textbf{value loss} (for the Critic) to evaluate behavior learning dynamics (Figure \ref{fig:loss_behavior}). Regarding the policy loss (Figure \ref{fig:loss_policy}), the KAN-based variants demonstrate a slower rate of descent compared to the baseline. More critically, the value loss trajectories (Figure \ref{fig:loss_value}) reveal a distinct performance gap: the baseline converges to a lower and more stable value (orange line), whereas the KAN and FastKAN variants stabilize at a higher loss level with visibly larger fluctuations.

\section{Discussion}
\label{Discussion}

In this section, we synthesize the experimental findings to evaluate the feasibility and implications of replacing specific DreamerV3 components with KAN-based architectures. The discussion is organized by functional modules, drawing evidence from the quantitative benchmarks (Table \ref{tab:component_ablation}), sample efficiency curves (Figure \ref{fig:training_dynamics_all}), wall-clock efficiency analysis (Figure \ref{fig:time_efficiency_all}), and component optimization dynamics (Figures \ref{fig:loss_image}--\ref{fig:loss_behavior}).

\subsection{Visual Perception}
\label{Discussion:Visual Perception}

\textbf{Sample Efficiency and Time-to-Solution.} 
First, regarding sample efficiency, observations from Table \ref{tab:ablation_vis} and Figure \ref{fig:vis_score} indicate that while the agent eventually learns a successful policy, the replacement of the visual encoder/decoder leads to a significant degradation in sample efficiency. Although the training throughput (FPS/Policy, FPS/Train) of KAN and FastKAN variants exceeds that of the CNN-based baseline (as shown in Table \ref{tab:ablation_vis} and Figure \ref{fig:vis_score_time}), the wall-clock time required to reach the high-performance threshold (reward $\ge 900$) remains considerably longer. This implies that the computational speedup provided by FastKAN is insufficient to offset the increased sample complexity. Consequently, we conclude that directly replacing the CNN backbone with fully connected KAN architectures is a suboptimal strategy for high-dimensional visual control tasks.

\textbf{Root Cause Analysis: The Absence of Spatial Inductive Bias.} 
To understand the underlying cause of this inefficiency, we examine the component optimization dynamics in Figure \ref{fig:loss_image}. The reconstruction loss for KAN-based autoencoders plateaus at a high value ($\approx 100$) and fails to converge to zero, unlike the CNN baseline. 
It is well-established that visual data exhibits strong \textbf{spatial locality}, where neighboring pixels share significant dependencies. Convolutional Neural Networks (CNNs) leverage this property through the convolution operator, introducing a critical \textbf{spatial inductive bias} (e.g., translation invariance and locality). This prior knowledge frees the network from learning local pixel correlations from scratch, drastically reducing the hypothesis space.
In contrast, our implementation flattens the image into a 1D vector before feeding it into KANs. Despite the theoretical function approximation capabilities of KANs \citep{liu2024kan} and the success of KAE \citep{yu2024kae} on simple datasets like MNIST, our results suggest that without structural priors, KANs struggle to capture the complex dependencies in high-resolution, high-dimensional observations (e.g., DMC). The persistent high reconstruction loss confirms that the model fails to learn an effective visual representation, which cascades into poor downstream policy learning.

\textbf{Throughput Trade-off.} 
Finally, we address the seemingly counter-intuitive observation in Table \ref{tab:ablation_vis} that KAN-based visual modules achieve higher training FPS than the baseline. While KANs are generally considered computationally heavier than MLPs, the comparison here is against \textbf{CNNs}. Convolutional operations, despite hardware acceleration, involve computationally intensive sliding window calculations and complex memory access patterns, which can be more expensive than the matrix-multiplication-based operations of FastKAN (once inputs are flattened). However, this throughput advantage is superficial. Given the poor sample efficiency, prolonged convergence time, and failure in reconstruction, the speed advantage of KANs in this context is rendered negligible. The lack of representation quality outweighs the benefits of faster gradient updates.

\subsection{Latent Prediction}
\label{Discussion:Latent Prediction}

\textbf{Sample Efficiency and Time-to-Solution.} 
First, regarding sample efficiency, data from Table \ref{tab:ablation_pred} and Figure \ref{fig:pred_score} demonstrate that both KAN and FastKAN variants achieve asymptotic returns comparable to the baseline. Notably, the FastKAN variant converges to the high-performance threshold with slightly fewer interaction steps than the baseline, suggesting a marginal improvement in sample efficiency. 
In terms of wall-clock training time (Figure \ref{fig:pred_score_time}), the standard KAN variant is noticeably slower due to its computational overhead. However, the FastKAN variant, despite a slightly lower interaction FPS than the MLP baseline, achieves a comparable (or even slightly shorter) total time-to-convergence. This efficiency gain is attributed to its improved sample efficiency offsetting the minor throughput reduction. Considering experimental stochasticity, we conclude that FastKAN serves as a robust \textbf{drop-in replacement} for MLPs in latent predictors without compromising—and potentially enhancing—overall training efficiency.

\textbf{Optimization Dynamics and Loss Interpretation.} 
Although Figure \ref{fig:loss_prediction} indicates that the baseline MLP achieves a faster initial reduction in training loss compared to FastKAN, this metric does not linearly correlate with downstream policy performance. First, RL training is inherently non-stationary; as the policy improves, the agent encounters novel states with differing reward dynamics. A persistently non-zero loss may indicate that the KAN-based predictor is actively adapting to these new, high-value regions rather than overfitting to early, trivial states. Furthermore, the scalar loss represents an average over the batch, which can be misleading in tasks with sparse rewards or class imbalance (e.g., Continue flags). A trivial predictor that outputs constant zeros often yields a lower average loss than an informative predictor that captures sparse high-value signals with some variance. The fact that FastKAN supports competitive policy performance (Figure \ref{fig:ac_score}) despite slightly higher prediction loss suggests it avoids such trivial solutions, effectively capturing the critical—albeit sparse—learning signals required for policy improvement.

\textbf{Structural Generalization via KA Theorem.} 
Crucially, training loss measures performance only on the replay buffer distribution, whereas the efficacy of a World Model depends on its generalization to unseen states during latent imagination. Theoretically, MLPs rely on the \textbf{Universal Approximation Theorem (UAT)}, which guarantees approximation capabilities but offers no guarantees on the functional form outside the training data. In contrast, KANs are grounded in the \textbf{Kolmogorov-Arnold representation theorem}, which posits that multivariate functions can be represented via the composition of learnable univariate functions. We hypothesize that if KAN-based predictors successfully learn the constituent sub-functions of the reward signal, the resulting "reward landscape" may be topologically more accurate than that of an MLP approximation. Consequently, KANs could offer superior \textbf{structural generalization} to unseen states, providing more reliable guidance for long-horizon planning, even if this structural precision is not fully reflected in the scalar training loss.

\subsection{Policy Optimization}
\label{Discussion:Policy Optimization}

\textbf{Sample Efficiency and Time-to-Solution.} 
Regarding sample efficiency, observations from Table \ref{tab:ablation_ac} and Figure \ref{fig:ac_score} mirror the findings in the Visual Perception experiments: replacing the Actor and Critic networks with KAN or FastKAN architectures leads to a marked degradation in sample efficiency. In terms of wall-clock training time (Figure \ref{fig:ac_score_time}), both variants are significantly slower than the baseline, with the standard KAN suffering from severe computational delays and FastKAN failing to compensate for the extra sample complexity with its throughput. Based on these dual deficits in data and time efficiency, we conclude that a direct, drop-in replacement of MLPs with current KAN architectures in the Actor-Critic loop is a suboptimal strategy for this benchmark.

\textbf{Optimization Dynamics and Loss Interpretation.} 
The component optimization curves in Figure \ref{fig:loss_behavior} provide proximate causes for this performance gap. The policy loss (Figure \ref{fig:loss_policy}) exhibits a sluggish rate of decay compared to the baseline, indicating inefficient policy improvement. More critically, the value loss (Figure \ref{fig:loss_value}) converges to a significantly higher value with larger variance. This implies that the KAN-based Critic struggles to accurately approximate the return distribution, which in turn destabilizes the Actor's learning process by providing noisy or biased advantage estimates.

\textbf{Task Complexity: Value Estimation vs. Reward Prediction.} 
We attribute these difficulties to the fundamental difference in task complexity between \textit{Reward Prediction} (Section \ref{Discussion:Latent Prediction}) and \textit{Value Estimation}. While the reward function often follows a deterministic, formulaic structure relative to the state (which KANs fit well), the Critic in DreamerV3 must approximate the \textbf{distribution of $\lambda$-returns}. This target involves \textbf{long-horizon temporal aggregation} of stochastic rewards and relies on \textbf{bootstrapping} from the Critic's own future predictions. Unlike the supervised regression of instantaneous rewards, value estimation is a recursive, non-stationary problem where errors compound over the imagination horizon. Our results suggest that standard KANs, without specific architectural priors, lack the optimization stability to efficiently approximate these complex, recursive value landscapes involving future uncertainty.

\subsection{Synthesis and Conclusion}

Synthesizing the above analysis, we derive three primary conclusions regarding the integration of KANs into World Models:

\begin{itemize}
    \item \textbf{Suboptimal for Visual Perception due to Missing Inductive Biases:} Direct replacement of the CNN-based Autoencoder with standard KANs or FastKANs proves to be a suboptimal strategy. The primary constraint is not a lack of expressivity, but rather the \textbf{absence of structural priors}. By flattening the image, KANs lose the crucial \textbf{spatial inductive bias} that CNNs inherently possess, making it inefficient for them to learn the complex local correlations in high-dimensional visual data from scratch.
    \item \textbf{Suboptimal for Policy Optimization due to Task Complexity:} Similarly, replacing the Actor-Critic MLP backbones is also suboptimal. While these components do not require spatial priors, they face a different challenge: approximating the \textbf{complex, high-dimensional, and recursive value landscapes} defined by the Bellman equation. Our results indicate that standard KANs struggle with the optimization stability required for this non-stationary task, leading to poor sample efficiency.
    \item \textbf{Promising for Latent Prediction:} Conversely, FastKAN emerges as a \textbf{robust and efficient drop-in replacement} for MLPs in the Reward and Continue predictors. These components require modeling deterministic, low-dimensional mappings—a task that is neither spatially structured nor recursively defined. This aligns perfectly with the mathematical strengths of KANs, allowing FastKAN to achieve competitive performance and training stability without the computational bottlenecks or inductive bias limitations observed elsewhere.
\end{itemize}

\section{Conclusion}
\label{Conclusion}

\textbf{Summary of Findings.} 
In this work, we presented \textbf{KAN-Dreamer}, an exploratory study evaluating the integration of KAN and FastKAN architectures into the DreamerV3 framework. Our experiments demonstrate that \textbf{FastKAN} exhibits significant potential as a drop-in replacement for MLPs in the \textbf{Reward and Continue Predictors}. Under iso-parameter conditions, FastKAN-based predictors achieve performance parity with MLPs in terms of asymptotic return, sample efficiency, and wall-clock training time. Furthermore, we extensively discussed the structural and optimization challenges that currently limit the applicability of standard KANs in high-dimensional Visual Encoders and complex Actor-Critic networks, providing a foundation for future architectural improvements.

\vspace{0.5em}

\textbf{Future Work.} 
Based on our findings, we identify some promising directions for future research:

\begin{enumerate}
    \item \textbf{Tailored Architectures and Hyperparameter Tuning:} 
    Our preliminary results suggest that a one-size-fits-all configuration for KANs is suboptimal. Future work will focus on co-optimizing both the \textbf{architecture} and \textbf{hyperparameters} for each specific component. On an architectural level, this involves exploring heterogeneous designs; for example, the visual encoder, processing high-frequency pixel data, may benefit from a finer grid granularity (higher capacity), whereas the reward predictor might be more stable with a coarser grid. On a hyperparameter level, this entails a systematic analysis of sensitivity to parameters like the RBF bandwidth or the input range clamping, tailored to the non-stationary data distribution of each subsystem in online reinforcement learning.

    \item \textbf{Interpretability and Symbolic Discovery:} 
    Leveraging the white-box nature of KANs, we aim to move beyond black-box prediction. A key direction is to utilize symbolic regression to \textbf{visualize and interpret the learned reward functions}. This, however, introduces a non-trivial architectural challenge stemming from a mismatch with DreamerV3's design. To enhance stability across diverse reward scales, DreamerV3 frames reward prediction not as a direct regression task, but as a \textbf{classification problem} over discrete bins using a two-hot loss formulation. To enable symbolic discovery of a reward function (e.g., $r = f(s_t)$), the KAN-based predictor head would likely need to be reverted to a \textbf{direct regression output}. A crucial avenue for future research is to investigate this modification, carefully balancing the pursuit of interpretability against a potential reduction in the training stability and generality that the original classification-based approach provides. Successfully navigating this trade-off could reveal intrinsic relationships between state factors in sparse-reward tasks, offering a transparent view of the agent's learned objectives.

    \item \textbf{Recurrent KAN Architectures for Temporal Modeling:} 
    Our current study treats KANs as stateless, feed-forward function approximators. A significant future direction is to explore architectures that directly embed temporal reasoning into the KAN framework itself, moving beyond simple drop-in replacement. This could involve two primary avenues: (1) designing truly \textbf{Recurrent KANs (KAN-RNNs)}, where the learnable activation functions on the edges are stateful or influence a recurrent hidden state, potentially offering an alternative to the GRU-based sequence model. (2) Investigating hybrid models where KANs are used to parameterize the transition functions \textit{within} the existing RSSM framework. The goal is to determine if KANs' superior function approximation capabilities can lead to more accurate and efficient long-horizon predictions in the latent space.

    \item \textbf{Incorporating Spatial Priors:} 
    To address the limitations in visual perception, we propose integrating \textbf{spatial inductive biases} into the KAN framework (e.g., Convolutional KANs). Furthermore, we intend to investigate advanced regularization techniques to constrain the expressivity of KANs, preventing overfitting to high-frequency noise in visual reconstruction tasks.
\end{enumerate}

\bibliographystyle{plainnat} 
\bibliography{references} 

\newpage


\appendix

\setcounter{table}{0}
\setcounter{figure}{0}
\renewcommand{\thetable}{A\arabic{table}}
\renewcommand{\thefigure}{A\arabic{figure}}
\renewcommand{\thesection}{Appendix \Alph{section}} 

\section{Detailed Architectural Specifications}
\label{app:architecture}

To ensure a fair comparison under the iso-parameter constraint ($\approx$ 10.5M total parameters), we adjusted the scaling hyperparameters for each component. Table \ref{tab:hyperparams_detail} details the specific configurations.

For the \textbf{Baseline} (DreamerV3), the visual encoder/decoder size is controlled by the \texttt{depth} parameter (which scales the channel multipliers), while vector-based components are controlled by \texttt{units}. For \textbf{KAN} and \textbf{FastKAN} variants, we adjust the number of \texttt{units} (hidden width) to match the parameter count. Notably, for FastKAN, we align the number of RBF centers with the grid size ($G$) of standard KANs to maintain architectural consistency.

\begin{table}[h]
    \centering
    \caption{\textbf{Detailed architectural configurations.} 
    \textbf{Scale Config:} Denotes the primary hyperparameter used to scale the model size (\texttt{depth} for CNNs, \texttt{units} for MLPs/KANs).
    \textbf{$G$:} Grid size (for KAN) or Number of RBF Centers (for FastKAN).
    \textbf{$k$:} Spline order (specific to KAN).}
    \label{tab:hyperparams_detail}
    
    \resizebox{\textwidth}{!}{%
    \begin{tabular}{l | l l c c | c}
        \toprule
        & \multicolumn{4}{c|}{\textbf{Component Configuration}} & \textbf{Total Model} \\
        \textbf{Model ID} & \textbf{Backbone} & \textbf{Scale Config (Depth/Units)} & \textbf{Grid Size ($G$)} & \textbf{Order ($k$)} & \textbf{Params} \\
        \midrule
        \multicolumn{6}{l}{\textit{\textbf{1. Perception Group} (Encoder \& Decoder)}} \\
        \midrule
        \textbf{Baseline} & CNN & \texttt{depth}: 16 & -- & -- & 10.49M \\
        KAN-Vis. & KAN & \texttt{units}: 8 & 8 & 3 & 10.40M \\
        FKAN-Vis. & FastKAN & \texttt{units}: 12 & 8 & -- & 10.51M \\
        \midrule
        \multicolumn{6}{l}{\textit{\textbf{2. Prediction Group} (Reward \& Continue Heads)}} \\
        \midrule
        \textbf{Baseline} & MLP & \texttt{units}: 256 & -- & -- & 10.49M \\
        KAN-Pred & KAN & \texttt{units}: 20 & 8 & 3 & 10.45M \\
        FKAN-Pred & FastKAN & \texttt{units}: 30 & 8 & -- & 10.51M \\
        \midrule
        \multicolumn{6}{l}{\textit{\textbf{3. Behavior Group} (Actor \& Critic)}} \\
        \midrule
        \textbf{Baseline} & MLP & \texttt{units}: 256 & -- & -- & 10.49M \\
        KAN-AC & KAN & \texttt{units}: 24 & 8 & 3 & 10.48M \\
        FKAN-AC & FastKAN & \texttt{units}: 34 & 8 & -- & 10.47M \\
        \bottomrule
    \end{tabular}%
    }
\end{table}

\newpage
\setcounter{table}{0}
\renewcommand{\thetable}{B\arabic{table}}

\section{Training Hyperparameters}
\label{app:hyperparams}

We adopt the standard hyperparameter settings from the official DreamerV3 implementation \citep{hafner2025mastering} across all experiments. Table \ref{tab:dreamer_hyperparams} summarizes the configuration used for the baseline and all KAN-based variants.

\begin{table}[h]
    \centering
    \caption{\textbf{DreamerV3 Hyperparameters.} These settings are kept consistent across all experiments to ensure a fair comparison. The notation $\text{Per}(R, p)$ denotes the $p$-th percentile of the return distribution.}
    \label{tab:dreamer_hyperparams}
    
    \small 
    \begin{tabular}{l c c}
        \toprule
        \textbf{Name} & \textbf{Symbol} & \textbf{Value} \\
        \midrule
        \multicolumn{3}{l}{\textit{\textbf{General}}} \\
        \midrule
        Replay capacity & --- & $5 \times 10^6$ \\
        Batch size & $B$ & 16 \\
        Batch length & $T$ & 64 \\
        Activation & --- & RMSNorm + SiLU \\
        Learning rate & --- & $4 \times 10^{-5}$ \\
        Gradient clipping & --- & AGC(0.3) \\
        Optimizer & --- & LaProp($\epsilon = 10^{-20}$) \\
        \midrule
        \multicolumn{3}{l}{\textit{\textbf{World Model}}} \\
        \midrule
        Reconstruction loss scale & $\beta_{\text{pred}}$ & 1 \\
        Dynamics loss scale & $\beta_{\text{dyn}}$ & 1 \\
        Representation loss scale & $\beta_{\text{rep}}$ & 0.1 \\
        Latent unimix & --- & 1\% \\
        Free nats & --- & 1 \\
        \midrule
        \multicolumn{3}{l}{\textit{\textbf{Actor Critic}}} \\
        \midrule
        Imagination horizon & $H$ & 15 \\
        Discount horizon & $1/(1 - \gamma)$ & 333 \\
        Return lambda & $\lambda$ & 0.95 \\
        Critic loss scale & $\beta_{\text{val}}$ & 1 \\
        Critic replay loss scale & $\beta_{\text{repval}}$ & 0.3 \\
        Critic EMA regularizer & --- & 1 \\
        Critic EMA decay & --- & 0.98 \\
        Actor loss scale & $\beta_{\text{pol}}$ & 1 \\
        Actor entropy regularizer & $\eta$ & $3 \times 10^{-4}$ \\
        Actor unimix & --- & 1\% \\
        Actor RetNorm scale & $S$ & $\text{Per}(R, 95) - \text{Per}(R, 5)$ \\
        Actor RetNorm limit & $L$ & 1 \\
        Actor RetNorm decay & --- & 0.99 \\
        \bottomrule
    \end{tabular}
\end{table}

\newpage 

\setcounter{figure}{0}
\renewcommand{\thefigure}{C\arabic{figure}}

\section{Additional Preliminary Results on Manipulation Tasks}
\label{app:metaworld}

To assess the generality of FastKAN-based predictors, which emerged as the most promising variant in our main experiments, we conducted preliminary experiments on the MetaWorld suite, specifically the \textit{Reach} task. 

\begin{figure}[h]
    \centering
    \includegraphics[width=\textwidth, trim=10 280 2 280, clip]{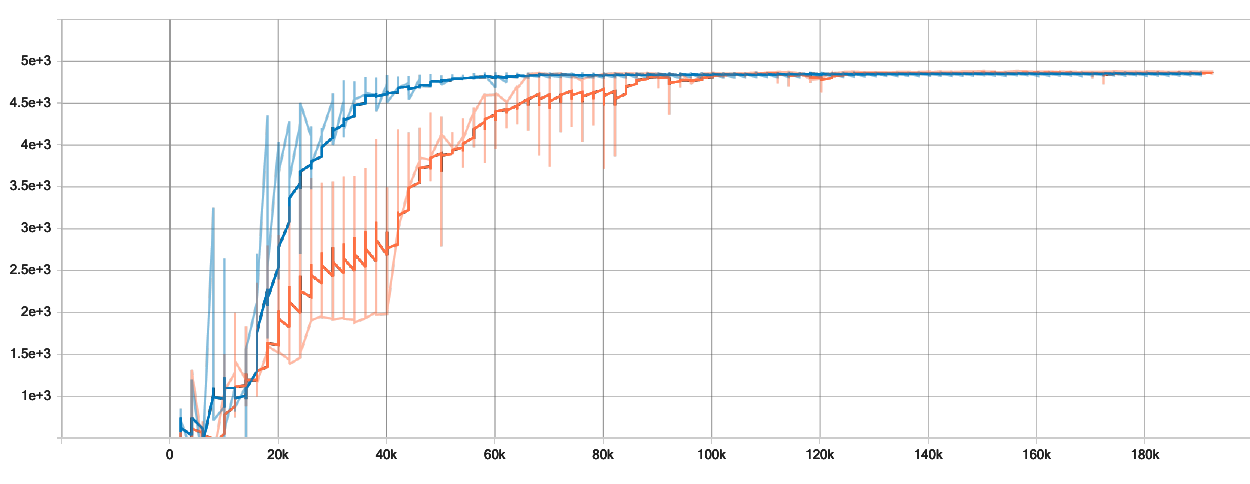}
    \caption{\textbf{Episode return on the MetaWorld Reach task.} Comparison of the FastKAN-based predictor (blue) against the MLP baseline (orange). The curves represent smoothed episode returns over environment steps.}
    \label{fig:metaworld_reach}
\end{figure}

Figure \ref{fig:metaworld_reach} illustrates the training trajectories of the FastKAN-based predictor compared to the MLP baseline. For this specific domain, we observe that the FastKAN backbone maintains numerical stability and exhibits competitive convergence properties. To achieve this, we performed a light hyperparameter sweep over the number of RBF centers ($G$) for the FastKAN model, while adjusting its hidden \texttt{units} to adhere to the iso-parameter constraint against the MLP baseline. However, extensive multi-seed evaluations were not the primary focus of this specific ablation study.

\end{document}